\documentclass[conferance,12pt,onecolumn,letterpaper,final]{IEEEtran}

\usepackage{graphicx}
\usepackage{float}
\graphicspath{{Figures/}}
\usepackage{hyperref}
\hypersetup{
    colorlinks=true,
    linkcolor=blue,
    filecolor=magenta,      
    urlcolor=cyan,
    pdftitle={Sharelatex Example},
    bookmarks=true,
    pdfpagemode=FullScreen,
}
\usepackage[numbers]{natbib}
\usepackage{fixltx2e}

\begin{document}

\title{Motion Prediction on Self-driving Cars: A Review}
\author{\IEEEauthorblockN{Shahrokh~Paravarzar\IEEEauthorrefmark{1} and
Belqes~Mohammad \IEEEauthorrefmark{2}}\\
\IEEEauthorblockA{University of Alberta, Department of Computer Science, Multimedia Research Lab\\
Email: \IEEEauthorrefmark{1} paravarz@ualberta.ca ,
\IEEEauthorrefmark{2} belqes@ualberta.ca }}

\maketitle
\thispagestyle{plain}
\pagestyle{plain}

\begin{abstract}
The autonomous vehicle motion prediction literature is reviewed. Motion prediction is the most challenging task in autonomous vehicles and self-drive cars. These challenges have been discussed. Later on, the state-of-the-art has reviewed based on the most recent literature and the current challenges are discussed. The state-of-the-art consists of classical and physical methods, deep learning networks, and reinforcement learning. prons and cons of the methods and gap of the research presented in this review. Finally, the literature surrounding object tracking and motion will be presented. As a result, deep reinforcement learning is the best candidate to tackle self-driving cars. 
\end{abstract}

\IEEEpeerreviewmaketitle

\section{Introduction}
\begin{figure}[H]
\begin{center}
\includegraphics[width=0.7\textwidth]{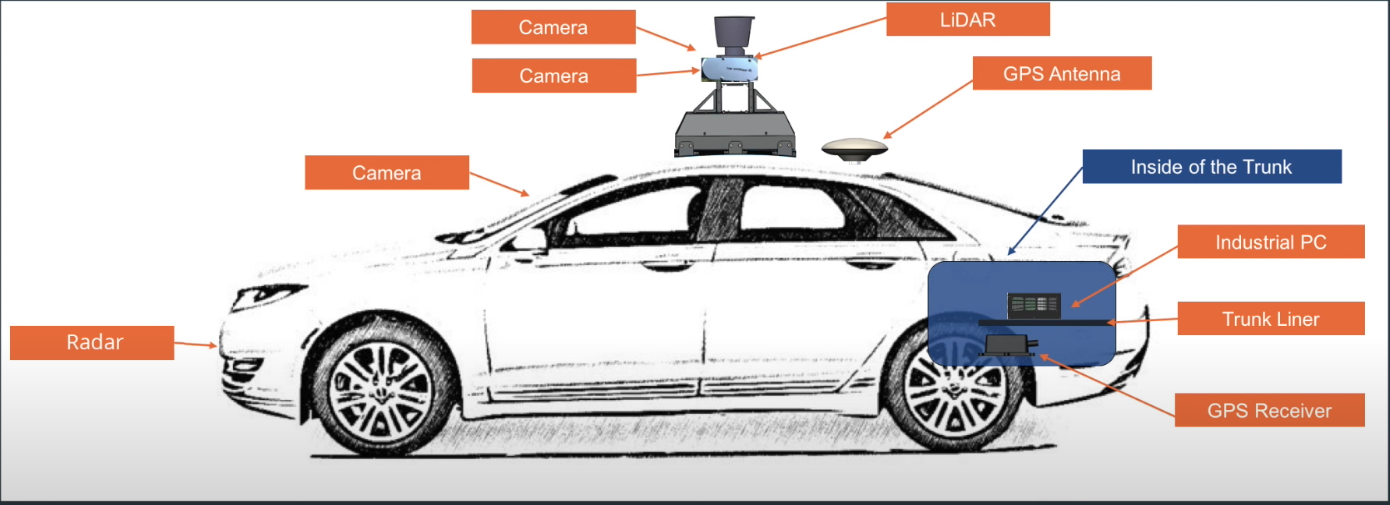}
\caption{Schematic representation of self-drive cars sensors}
\label{SDC-Structure}
\end{center}
\end{figure}

\IEEEPARstart{A}{}Autonomous vehicles and self-drive cars will be part of our life and community in the near future. The process for design autonomous self-drive cars is a complex task. The automation process of self-drive cars is started by generating high-quality maps from the driving environment such as buildings roads and etc. These maps are called "HD maps" or "semantic maps". The process is continued by localization. Self-driving cars should find and recognaize their exact location in the real world. The next step is perception which will enable self-driving cars to sense their surroundings such as lanes, other cars, traffic signs and etc. After this step, the self-driving car should predict the next movement or trajectories, based on the action of surrounding agents such as other vehicles, pedestrians, and etc. This step is called a "prediction". Having known the upcoming event, self-driving vehicle will be planning the future trajectory. To accomplish the planned trajectories, self-driving cars will use the control algorithms. All these steps should result in safe travel under the high safety and security environment. Figure~\ref{SDC-Structure} shows the schematic representation of these steps. Among all the steps motion prediction has got a unique complexity due to additional interaction with the environment. Prior to the discussion about this topic, it would be useful to clarify the technical jargons which will be used in the following sections. 

\section{Basics and Challenges of Vehicle Motion Prediction}
After the localization and perception process for the self-drive car (SDC), SDC requires to consider possible future scenarios based on its current state. All the future possible scenarios rely on the behavior of the surrounding environment and the new information which will be provided by the sensors as a planning step. Motion prediction plays a significant role in the autonomous driving application as it is a support for decision making~\cite{lefevre2014survey} and enables us to assess the risks associated with the planning process~\cite{zhan2018probabilistic}. In the literature, the words behavior, motion, and maneuver are used interchangeably. In the following section, the challenges with motion prediction will be discussed. After that, the terminology will be defined and finally, the probabilistic representation of the problem will be identified.
\subsection{Technical Concepts and Terminology Definitions}
To define the problem of the self-driving vehicle and smart vehicle, it is useful to define the technical terminologies in this area
\begin{itemize}
\item \textbf{\small Semantic Map} Every type of modeling for self- driving cars requires high-resolution maps. The semantic maps are crucial for the autonomous vehicle (AV) from an operational and safety point of view. The semantic maps are providing a better understanding of AV's from their surrounding environment. The semantic maps consist of several layers such as road networks, building in 3D, lane layer, pathways and etc.
\item \textbf{\small Sensors} There are three major sensors installed on self-drive cars, cameras, Radar (Radio Detection and Ranging), and LiDAR (Light Detection and Ranging). The cameras are available to display highly detailed and realistic images from the surrounding environment. This would enable the self-drive cars to detect the objects and classify them. The radar sensor is working based on the doppler effect. There exist two types of radar sensors, long- and short-range. The long-range radar sensors (77 GHz) are controlled the automatic distance control and break distance control. The short-range (24 GHz) enables blind-spot monitoring. By LiDAR it is possible to localize the self-drive cars by means of point cloud matching. The LiDAR sensor could provide a fully 360-degree map around the vehicle. This sensor detects the data from the current environment and matches it with pre-existing data in a high definition map. This approach will yield the global position and heading the car on a semantic map. There are different algorithms to match the point cloud with a high definition map such as an interactive closest point (ICP), Filter approach, Histogram Filter, Kalman Filter \cite{LyftLevel5}. 
\item \textbf{\small Target Vehicles (TVs)} are those vehicles which their behavior is important to be predicted by the model and the self-drive car is in interaction with them. 
\item \textbf{\small Ego Vehicle (EV)} a self-drive car which is moving while observing the behavior of the TVs
\item \textbf{\small Surrounding Vehicles (SVs)} are whose behavior impacts the self-driving car. Choosing the SVs are adapting with different algorithms and depends on the model assumptions Non-Effective Vehicles (NVs)~\cite{mozaffari2020deep}.
\end{itemize}
\section{Challenges}
Vehicles are the non-holonomic system that can not change their trajectories instantly upon the new status. This makes the vehicle's movement prediction challenging. The intrinsic behavior of a vehicle can be controlled with two parameters of acceleration and break. In normal life, the decision about these parameters should be identified based on a human decision. The human decision is a reflection of the vehicle environment, however, the decisions for the AV's will be taken by the algorithms. This issue is a challenging task due to the following reasons:  
\begin{itemize}
\item In real scenarios, the ovehicles's behavior on the road will affect the behavior of the other vehicles, and there existed an interdependent relation with respect to their status. The sudden maneuvers, lane changes, stops on the road would affect the movement and trajectory of the other vehicles. For this reason, the prediction of the behavior of the other vehicles and its effect on the EV is a complex task and challenging. 
\item The traffic rules, signs, and geometry of the road has a direct effect on the behavior of vehicles on the road. In addition to this, sudden changes in the road and environmental conditions or changes in the climate will affect the vehicle's behavior on the road. 
\item There are several possible scenarios for the vehicles in different situations. This behavior of the AV is called multimodal behavior. The models provide comprehensive behavior prediction outcome for the behavior prediction to provide reliable behavior for the vehicle on the road.
\end{itemize}
In addition to the above-mentioned challenges, the decision for this type of problem should be taken less than a second. This issue is another challenge because the computational capacity on-board is restricted in autonomous vehicles. 
\section{Mathematical Representation of the Problem}
The motion prediction problem has a probabilistic nature. It is associated with uncertainty and risk evaluation is a key feature of this problem. The future behavior of the AV can be presented as (Eq.~\ref{eq:1}): 
\begin{equation}
\label{eq:1}
{\chi}_{TVs} = \{ x_{t}^{i} , x_{t+1}^{i}, \dots , x_{t+m}^{i}\}_{i=1}^{N} 
\end{equation}
Where ${x}_{i}^{t}$ represents the states (e.g., position) of vehicle $i$ at time step $t$, $N$ is the number of TVs, and $m$ is the length of
the prediction window. The goal is to calculate the conditional distribution of $p ({\chi}_{TVs} | {O}_{EV})$ where ${O}_{EV}$ are the available
observations to the EV. To calculate the conditional probability many researchers drop the interaction between the vehicles. Consequently, the conditional probability of just one vehicle will be calculated in each step~\cite{mozaffari2020deep}. 
\section{Classification of Existing Works}
\citet{lefevre2014survey} classified the prediction model to (i) physics-based, (ii) maneuver-based, and (iii), interaction-aware model.~\citet{mozaffari2020deep} proposed the state-of-the-art and classified them based on the (i) input representation, (ii) output type,  and (iii) prediction method.~\citet{kiran2020deep} reviewed the reinforcement learning approach for autonomous vehicle. In this review, our major focus will be on deep learning and reinforcement learning approach.
\subsection{Deep learning approach for Autonomous Vehicles}
The deep learning methods for AV behavior prediction can be categories into three classes (i) recurrent neural network, (ii) convolutional neural network, and (iii) other methods (e.g. combined methods or graph neural networks). 
Gated Recurrent Unit (GRU)~\cite{cho2014learning} and long short-term memory (LSTM)~\cite{hochreiter1997long} are two of the main architecture for behavior prediction.\\
~\citet{altche2017lstm} are trained and validated the LSTM network to predict future longitudinal and lateral trajectories for vehicles on the highway using the NGSIM dataset. Their network is showed better accuracy with respect to the previous state-of-the-art. the average RMS error for the lateral position was 70 $cm$ and 10 $s$ in the future and lower than 3${ms}^{-1}$ for longitudinal velocity with the same horizon. Their model output provides a single value for the trajectory prediction, however, they suggested the probability-based model for future work. The advantage of their model can be summarized in its generalization considering 6000 vehicles at the same time. \\
\citet{xin2018intention} are presented the long-horizon trajectory prediction of surrounding vehicles using a dual long short-term memory to predict the interaction of EV with SVs. The first LSTM block recognizes the driver's intention as an indicator and the LSTM will predict the future trajectory. The dual LSTM model provided the future trajectories over 5$s$ time with a mean square error (RMSE) equal to 0.577$m$ for longitudinal and 0.49$m$ for lateral prediction.Their work provided small prediction bounds for the lateral position which would lead to the adaptability of the model to various road geometry. They suggested future research by generalizing the model for different scenarios such as intersections and unstructured roads and adding the stochasticity to the model output.\\
\citet{DBLP} presented a multiple-layer LSTM model to provide an interaction aware prediction of surrounding vehicles. The model provides a multi-modal distribution over future motion using the NGSIM US-101 and I-80 dataset. \citet{DBLP} used six different decoder LSTMs which correlate with six specific manoeuvres of highway driving. An encoder LSTM is applied to the past trajectory of vehicles. The hidden state of each decoder LSTM is initialized with the concatenation of the last hidden state of the encoder LSTM and a one-hot vector representing the manoeuvres specific to each decoder. The decoder LSTMs predict the parameters of manoeuvres conditioned bivariate Gaussian distribution of future locations of the TV. Another encoder LSTM is also used to predict the probability of each of six manoeuvres.\\
\citet{zhao2019multi} implemented the combination of LSTM and CNN methods by concatenating the vector of the agent's movement and status scene encoded by LSTM and CNNs and respectively fed to U-net like network. The input and output of the mentioned network fed to the LSTM encoder to predict the future trajectory for the agents. \\
\citet{lee2017convolution} suggested the combination of six convolutional neural networks to predict the lane change intention and predictive control. They used the simplified bird's-eye view to lower the computational cost. These types of CNN models are not able to model the data series.\\
\citet{Yeping2018} introduced the Semantic-based Intention and Motion Prediction(SIMP) method to predict the different possible scenarios by using a fully connected neural network. They concluded that using the semantics in a single framework while combining several tasks would result in competitive results in comparison to traditional approaches.\\
\citet{casas2018intentnet} presented IntentNet as a learnable end-to-end model that combines the semantic high-resolution map with 3D point clouds produced by LiDAR sensors. They proved that how more complicated algorithms can model statistical dependencies between discrete and continuous intention. There is a chance to extend their model for pedestrian and bicycle intention. 
\subsection{Reinforcement learning approach for Autonomous Vehicles}
Reinforcement learning is an approach that an agent will learn by interacting with its environment by experiencing several scenarios. An agent improves its performance by maximizing the cumulative rewards received over its lifetime. The agent gradually improves its knowledge for the long-term by exploring the new knowledge and exporting the previous knowledge. The challenge in reinforcement learning is to control the trade off between exploration and exploitation. Doing that, a self-drive car would predict the future trajectories by experiencing different scenarios and improve its reaction with respect to its environment. The methods for predicting the trajectory of the self-drive car can be classified into three class of (i) Value-based methods (ii) Policy-based methods (iii) Actor-critic methods (iv) Model-based (vs. Model-free) \& On/Off Policy methods and (v) Deep reinforcement learning.\\
Autonomous driving tasks where RL could be applied include: controller optimization, path planning, and trajectory optimization, motion planning, and dynamic path planning, development of high-level driving policies for complex navigation tasks, scenario-based policy learning for highways, intersections merge and split, reward learning with inverse reinforcement learning from expert data for intent prediction for traffic actors such as pedestrian, vehicles and finally learning of policies that ensure the safety and perform risk estimation.\\
A comprehensive review on the different state and action representations which are used in autonomous driving research can be find in~\citet{leurent2018survey}. Commonly used state space features for an autonomous vehicle include: position, heading and velocity of ego-vehicle, as well as other obstacles in the sensor view extent of the ego-vehicle.\\
\citet{Keselman2018} provided a model-based reinforcement learning to dynamically plan the trajectories to provide a smooth control behavior of the vehicles. The new algorithm can be combined with trees and learnable heuristic from $A^{*}$ algorithm using a DQN over image-based input obstacle map. \\
\citet{ngai2011multiple} purposed a multiple-goal reinforcement learning (MGRL) method to solve the over-taking problem and avoid a collision between the vehicles. They used Q-learning (QL) or double-action QL to determine the action decision based to figure out if the agent and other vehicles would attempt to reach the same goal.\\
\citet{Isele2017} is provided an efficient strategy to tackle the intersection problem for self-drive cars in an occluded intersection and merge in to high way. Deep reinforcement learning is applied to find an optimal driving policy using LSTM for producing an internal state containing historical driving information and DQN for Q-function approximation.\\
\citet{Wang2018} purposed a DRL to solve the lane changing problem for autonomous vehicles using a Q learning. The method is based on approximation of a Q-function under having a continuous state space and action space assumption. Their results are shown that the DRL model could provide a smooth and efficient driving policy for lane-change maneuvers under an interactive driving environment. As future research, they suggested improving the model by providing the traffic rules and different road geometry. In addition to this, they suggested the combination of RL and MPC to make the best of both approaches.\\
\citet{Sallab_2017} provided a DRL framework by the combination of Q-network, RNN/LSTM, CNN/DNN using temporal and spatial aggregation. Finally, they test their model in the Open-source Racing Car Simulator (Torcs). 
\section{Gap of the Knowledge and Suggestions for Future Research}
A wide range of algorithms with a different resolution was introduced to the area of autonomous vehicles. The models are presented in a variety of resolutions, from a basic physical model for the various implementation of deep reinforcement learning. 
The suggestion for future research can be summarized as follows:
\begin{enumerate}
\item There is not any benchmark available to evaluate the performance of deep learning and reinforcement learning approach. As future work, a benchmark can be defined for behavior prediction which will evaluate the performance of several algorithms.
\item Most of the works consider that the AV has a full view of the surrounding environment which is not practical in real scenarios. In practice, there is not any sensor that provides a top-down view of an AV car. In addition to this,  sensor impairment would cause a problem for AV operation. 
\item The visual-auditory information such as signaling light or vehicle horn can be used to predict the future trajectory of EGV or SV which does not considered these interactions in their model.
\item The simulation environment and large training dataset may provide a proper reflection of the real world but the distribution of the events would be different in the reality. This would cause inaccuracy in the model. The portion and number of object classes in the simulation environment could play a significant role in model accuracy.
\end{enumerate}
\section{Miscellaneous Literature  - Motion and Tracking}
Tracking has two approaches, which are radically different. There were recognition-based  tracking and motion-based tracking. The work~\cite{murray1994motion} of Aloimonos Tsakiris, Bray, Wilco et al., Schalkoff and McVey, are examples of recognition-based systems. Recognition-based monitoring is just a refinement of the object recognition. In successive images, the object is recognized, and its location is retrieved. This monitoring system 's advantage is that it could be applied in three dimensions. They are also able to estimate the object's translation and rotation.\\
The significant limitation is that only a recognizable entity can be monitored. Recognition of objects is a mechanism at a high level that can be costly to build. Therefore, the efficacy of the monitoring system is limited by the effectiveness of a process of identification, as well as the types of objects detectable. Motion-based tracking systems vary greatly in that the moving target is solely detected through motion detection from recognition-based systems. They also have the advantage of being able to detect any moving object, regardless of size or form, and are thus more appropriate for our systems. It is essential to further subdivide motion-based techniques into motion-energy and optical flow tracking approaches.\\
In optically motion detection~\cite{sun2017smart} methods of optically surgical motion detection range from raw video processing to the use of specially developed motion recognition systems that use optical markers. These devices will not interrupt normal hand movements by using a camera or a set of cameras to assess the location of the object or hand. However, to find the right balance between occlusion and interference, they have to be correctly located.\\
In~\cite{sun2017smart} Glarner and colleagues performed a theoretical study, in which they used an operating room recording digital video analytics system. If the recording is done, an observer identifies a region of interest (ROI) in the footage. The template-matching tracking algorithm thereby implements the given ROI and produces kinematic data including velocity, displacement, and acceleration. Our technique is used for all motion data, regardless of the type of acquisition system used, because we strive to measure the motion data effectively and evaluate its performance. Ideally, in the acquisition system, it should not be bulky or hinder normal hand motions. The system should also be able to deploy safely in an aseptic environment. Thus, in our case study, we used a Leap Motion Controller as the motion detection method.
In the Mechanically-based motion detection establish hand and joint position data and also velocity data, gloves with embedded sensors may be used by a surgeon. The mechanical deformation of 'bend sensors' that translate hand motion into digital signals and adjustments in joint motion is measured by these devices. The mechanical deformation of 'bend sensors' that translate hand motion into digital signals and adjustments in joint motion is measured by these devices.\\
In the work~\cite{sun2017smart}  Electromagnetic-based motion detection, the Imperial College Surgical Assessment Device (ICSAD) was the commonly known scheme for collecting motion data during open surgery. Similar to magnetically based devices, this conventional method of electromagnetic tracking requires sensors to be placed on the hands and fingertips, preventing their use in a controlled operations center. Over the years, EM tracking technology has developed quickly. With an outside diameter of 1.5 mm, the magnetic sensor units are small. Compared to bulky mechanically derived glove systems, sensor embedded gloves that utilize this technology will do little to obstruct natural motion. The limitation is that it takes gloves customized with sensors. Removal of the gloves after surgery means the removal of the sensors, which can be expensive.\\
In~\cite{firouzmanesh2011perceptually,firouzmanesh2013perceptually} PCA-Based Approach Arikan suggested a way to compact Databases of large motion capture results, using a 30:1 to 35:1 ratio, with limited degradation of perceptual performance. The input data contains a file which defines the skeleton structure of the animated character, including a number of movement clips. In their studies, they used two databases. A first contained 620 K frames collected at 120 Hz (1:30 hour length). Although high compression ratios are obtained by this method, it may not be ideal for other applications. Second, the PCA-based approach achieves high compression ratios by exploiting similarities between DOFs, in contrast to temporal dependencies in each DOF. It might be ideal to retain the same compression ratio, though. Even if they encode each channel separately.\\
In ~\cite{firouzmanesh2011perceptually} implemented the optimal wavelet parameter choice for movement compression techniques. Their approach uses a distortion metric which calculates the difference both during compression between the positions within each joint and gives greater weight to joints with more local effect. For each channel, according to the desired compression ratio, an optimum number of wavelet coefficients is chosen such that distortion is reduced.\\
In MPEG-4 Bone-Based Animation, the strategy has a variety of drawbacks though delivering positive outcomes. The first, compression is very time-consuming due to the process of distributing an optimum number of parameters to each channel. Second, it is heavily contingent on the joint hierarchy that the error metric used for optimization.  In other words, assigning the joints with elevated failure rates more weight not inherently produce positive performance.\\
\citet{firouzmanesh2011perceptually} suggested two main techniques for encoding and distributing BBA. First approach is based upon predictive coding, followed by quantization and entropy coding. The second is the application for 1-D DCT to each DOF in the skeleton, followed by quantization and entropy coding. They also proposed a frame reduction technique to achieve greater compression.\\
In \cite{elnagar1995motion} the first trial to sense motion when the camera and various static objects are in motion is known as the earliest work of Ullman. He recommended a method of extraction of the corresponding optical flow into sets equivalent with each moving object in an image. Eventually, in the case of a transforming camera, Jain.t1 investigated the issue. After implementing a complicated logarithmic mapping on an image, they have used motion epipolar restraint.\\
The first to verbalize \cite{brown1987texture} the concept that surface orientation can be extracted from deviations in texture was Gibson [19501]. This problem has been discussed by many researchers since 1976 (cf. Aloimonos and Swain (1985) for a brief catalogue) using different texture and imaging geometry theories. The observations are that texture components, or texels, are dispersed over the surface within work. The consistency of their transmission is not a problem, nor is their shape. There would be an assumption that the variation in depth over the surface is small relative to the camera scale. Finally, Ohta (1981) 's perspective calculation is kept, and this also amounts to a limit on the size of texels relative to the surface curvature. We construct a constraint that derives purely from the geometry of imaging that extends to the area of imaged texel orientation. This limitation suggests that the form of the Lambertian surface illumination restriction is of a certain kind.\\
\citet{yin1999integrating} presented a system to track a talking face observed with an active camera for an arbitrary background, in the situation of a movable camera, the difficulties increases for searching and matching feature points in successive images since the background viewed is dynamically changing. The motion-energy tracking method can segment an image into regions of motion and in-activity by calculating the temporal derivative of an image sequence and thresholding at a suitable level to filter out the noise. This method is relatively simple and efficient. However, it is not suitable for application on an active camera system without modification. Since the active camera system can induce apparent motion on the scenes they view, compensation for camera motion must be made before the motion-energy detection technique can be used.\\
\citet{basu1990approximate} presented an algorithm to find the desired path, when one exists, with a required probability as a solution for the problem of finding a collision-free path connecting two points (start and goal) in the presence of obstacles, with constraints on the curvature of the path, is examined. This problem of curvature-constrained motion planning arises when (for example) a vehicle with constraints on its steering mechanism needs to be maneuvered through obstacles.\\
\citet{basu1987robust} presented a method for determining the translation of rigidly moving surface without correspondence for robotics applications, this method consists of four cameras that are used to recover the three translation parameters, instead of only the direction of translation ,in order to avoid point correspondences. It is used to deal with low levels of noise and has good behavior when the noise increases.\\
Regards to motion capture compression method,  presented technique of combing two (spatial segmentation and temporal blending) to achieve high compression rates and rendered quality on rhythmic motion data. motion sequences can be recorded using optical, mechanical or magnetic devices by tracking the movements of key points (such as joints) on an object. Human motion data show a considerable amount of spatial and temporal correlation, especially in rhythmic procedural movements, which can be exploited for effective compression.\\
Regards to motion capture compression method, \citet{firouzmanesh2013efficient}presented technique of combing two (spatial segmentation and temporal blending) to achieve high compression rates and rendered quality on rhythmic motion data. motion sequences can be recorded using optical, mechanical or magnetic devices by tracking the movements of key points (such as joints) on an object. Human motion data show a considerable amount of spatial and temporal correlation, especially in rhythmic procedural movements, which can be exploited for effective compression. 
\section{Conclusion}
A brief review of the different approaches is provided in this review for the issue of motion (behavior) prediction of autonomous vehicles. There exist two main approaches to solve the problem. The first category of the models is deep learning methods. A combination of CNN and RNN models are available in the literature to attack the problem on hand. The second category of the models belongs to reinforcement learning methods. This category seems to provide a more promising approach to deal with a problem. Researchers tried to provide more sophisticated models using reinforcement learning by adding several restrictions and driving rules to the models.  Although some of the models are considered a wide range of scenarios but still there is a gap in accuracy and complicity of the models to satisfy the level of safety that is required for the implementation of self-driving cars.

\bibliographystyle{IEEEtranN}
\bibliography{Bib.bib}


\end{document}